%% file: main.tex
\documentclass[10pt,twocolumn,letterpaper]{article}

\usepackage[pagenumbers]{iccv} 
\usepackage{lipsum}

\definecolor{iccvblue}{rgb}{0.21,0.49,0.74}
\usepackage[pagebackref,breaklinks,colorlinks,allcolors=iccvblue]{hyperref}
\def\confName{ICCV}
\def\confYear{2025}

\input{srcs/macros}

\title{RomanTex: Decoupling 3D-aware Rotary Positional Embedded Multi-Attention Network for Texture Synthesis}

\begin{document}
\twocolumn[{%
\renewcommand\twocolumn[1][]{#1}%

\author{
Yifei Feng\textsuperscript{1$*$} \quad
Mingxin Yang\textsuperscript{1$*$} \quad
Shuhui Yang\textsuperscript{1$*\dagger$} \quad
Sheng Zhang\textsuperscript{1} \quad
Jiaao Yu\textsuperscript{1} \quad
Zibo Zhao\textsuperscript{2} \\
Yuhong Liu\textsuperscript{1} \quad
Jie Jiang\textsuperscript{1} \quad
Chunchao Guo\textsuperscript{1$\ddagger$} \quad
\\[0.5em]
\textsuperscript{1}Tencent Hunyuan \ \ \ 
\textsuperscript{2} ShanghaiTech University\\
\href{https://oakshy.github.io/RomanTex/}{https://oakshy.github.io/RomanTex/}
}

\maketitle
\begin{center}
    \centering
    \captionsetup{type=figure}
    \includegraphics[width=\linewidth]{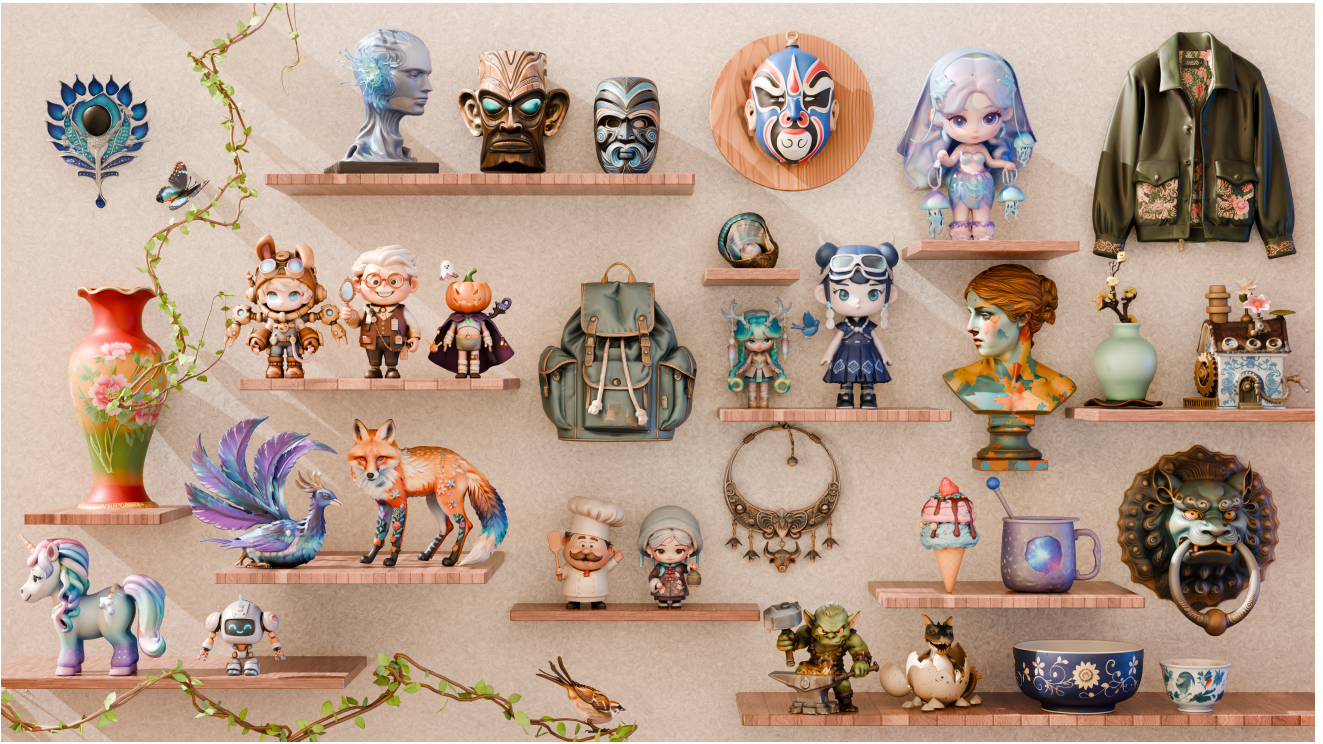}
    \captionof{figure}{3D assets with high quality textures generated by our method.}
    \label{fig:teaser}
\end{center}%
}]

\def\thefootnote{}\footnotetext{$*$ *Equal Contribution. $\dagger$ Project leader. $\ddagger$ Corresponding author.}

\input{srcs/0_abstract}
\input{srcs/1_intro}

\input{srcs/2_relatedwork}
\input{srcs/3_method}
\input{srcs/4_experiment}

\input{srcs/5_conclusion}

{\small
\bibliographystyle{ieeenat_fullname}
\bibliography{main}
}

\input{srcs/appendix}

\end{document}

%% file: srcs/macros.tex
\usepackage{graphicx}	
\usepackage{amsmath}	
\usepackage{amssymb}	
\usepackage{booktabs}
\usepackage{times}
\usepackage{microtype}
\usepackage{epsfig}
\usepackage{caption}
\usepackage{float}
\usepackage{placeins}
\usepackage{color, colortbl}
\usepackage{stfloats}
\usepackage{enumitem}
\usepackage{tabularx}
\usepackage{xstring}
\usepackage{multirow}
\usepackage{xspace}
\usepackage{url}
\usepackage{subcaption}
\usepackage{xcolor}
\usepackage[hang,flushmargin]{footmisc}
\usepackage{bm}

\usepackage{color}

\renewcommand{\thefootnote}{\fnsymbol{footnote}}

\newcommand{\ignore}[1]{}

\usepackage{booktabs}
\usepackage{siunitx}

%% file: srcs/0_abstract.tex
\begin{abstract}
Painting textures for existing geometries is a critical yet labor-intensive process in 3D asset generation. Recent advancements in text-to-image (T2I) models have led to significant progress in texture generation. Most existing research approaches this task by first generating images in 2D spaces using image diffusion models, followed by a texture baking process to achieve UV texture.  However, these methods often struggle to produce high-quality textures due to inconsistencies among the generated multi-view images, resulting in seams and ghosting artifacts.
In contrast, 3D-based texture synthesis methods aim to address these inconsistencies, but they often neglect 2D diffusion model priors, making them challenging to apply to real-world objects
To overcome these limitations, we propose RomanTex, a multiview-based texture generation framework that integrates a multi-attention network with an underlying 3D representation, facilitated by our novel 3D-aware Rotary Positional Embedding. Additionally, we incorporate a decoupling characteristic in the multi-attention block to enhance the model's robustness in image-to-texture task, enabling semantically-correct back-view synthesis.
Furthermore, we introduce a geometry-related Classifier-Free Guidance (CFG) mechanism to further improve the alignment with both geometries and images.
Quantitative and qualitative evaluations, along with comprehensive user studies, demonstrate that our method achieves state-of-the-art results in texture quality and consistency.
\end{abstract}

%% file: srcs/1_intro.tex
\section{Introduction}
\label{sec:intro}
Benefitting from the rapid advancements in large language models (LLM) and text-to-image (T2I) models, the integration of AI and 3D technology is optimizing the labor intensive traditional 3D pipeline in various ways, leading to a new wave of enthusiasm in the 3D industry at both software and hardware levels. Among these advancements, the generation of 3D assets is particularly transformative, encompassing both geometry generation and texture generation. Notably, texture synthesis has been a cardinal step of 3D content production. As Pixar co-founder Edwin Catmull states in \textit{Creativity, Inc.}~\cite{catmull2014creativity}: The reality of the digital world doesn’t come from perfect geometry; it comes from the imperfections on the surfaces. This pursuit of surface details is the core value of 3D asset texture generation, which plays a pivotal role in conferring both realistic material properties and exceptional artistic expressiveness to digital assets.

However, generating high-quality textures still remains a significant challenge, due to geometric complexity and the scarcity of 3D textured data. This motivates mainstream texture generation methods to draw inspiration from image generation models, which integrate geometric conditions with T2I models, thereby bridging the inherent dimensional gap between three-dimensional geometry and two-dimensional rendering spaces. Although this approach has achieved considerable success, it remains constrained by several critical limitations.

\textbf{Global consistency}: Considering the limited amount of 3D texture data, mainstream approaches predominantly leverage pre-trained T2I models. These methods either employ progressive inpainting~\cite{huang2024material,wu2024texro,zhang2024mapa,ceylan2024matatlas,zeng2024paint3d,chen2023text2tex,richardson2023texture} to fill in masked areas based on existing textures from a given viewpoint or utilize synchronous denoising~\cite{liu2025vcd,gao2024genesistex,liu2024text,zhang2024texpainter,cheng2024mvpaintsynchronizedmultiviewdiffusion} during the inference phase to achieve seamless texture generation. However, due to the lack of viewpoint awareness and 3D global perception in pre-trained models, these approaches typically result in global inconsistencies, which manifest as variations in texture styles and colors on the back of objects, furthermore the Janus problem. 

\textbf{Multi-view consistency}: Recently, multi-view diffusion models~\cite{shi2023zero123++,shimvdream,wang2023imagedream, Tang2023mvdiffusion,tang2025mvdiffusion++} leverage cross-view self-attention mechanisms to model correlations between multiple viewpoints, offering new insights for texture generation tasks. By fine-tuning pre-trained T2I models with 3D texture data, these approaches can provide multi-view images during the inference phase. Combined with the injection of geometric conditions, texture generation based on multi-view projections can be enabled~\cite{bensadoun2024meta, cheng2024mvpaintsynchronizedmultiviewdiffusion, vainer2024jointlygeneratingmultiviewconsistent, hunyuan3d2025}. Although the spatial consistency priors provided by 3D data can enhance the global consistency of semantics and the plausibility of rear textures, but merely relying on the soft constraints of cross-view attention mechanisms cannot guarantee highly consistent overlapping regions across views. Meanwhile, they still face challenges in balancing generalization and view consistency. Specifically, excessive training can lead to catastrophic forgetting, resulting in the loss of 2D priors and a decrease in texture diversity. Conversely, fine-tuning with a limited quantity of of parameters~\cite{huang2024mv} may struggle to ensure multi-view consistency.

\textbf{Image and geometric alignment}: Existing approaches generally utilize T2I-Adapter~\cite{mou2023t2iadapterlearningadaptersdig} or ControlNet~\cite{zhang2023adding} with depth and normal conditions for the injection of geometric information. However, the inherent complexity of image-to-texture tasks is exacerbated by the dual-modality conditioning paradigm, where concurrent utilization of image and geometric guidance introduces conditional conflicts—particularly pronounced in geometry-texture disentangled 3D generation frameworks~\cite{zhang2024claycontrollablelargescalegenerative, zhao2025hunyuan3d} This discrepancy arises from misalignments between generated geometries and reference images, effectively transforming texture synthesis into an ill-posed problem. Specifically, the network struggles to determine whether the synthesized textures should adhere to image guidance or geometric constraints. The former, such as IP-Adapter based image-to-texture method~\cite{jiang2024flexitex,sairaj_sig24} may result in insufficient image fidelity, while the later risks generating textures that fail to align with geometric edges, leading to textures being incorrectly baked onto the geometric surfaces during the texture mapping phase, resulting in color bleeding artifacts.

To address these challenges, we propose RomanTex, a novel texture generation framework based on a geometry-aware multi-view diffusion model. We introduce 3D-aware RoPE that allows the position map to bypass compression through a VAE encoder while mapping 3D consistent spatial coordinate information into a high-dimensional space. This facilitates the network's learning of geometric features and obviously alleviates cross-view consistency. Meanwhile, the decoupled multi-attention network architecture is designed to enable us to perform dropout training strategy of each attention module, allowing each component separately focus on generative fidelity, consistency and diversity. Additionly, we propose geometry-related CFG technique,  that balances adherence to reference images and geometric constraints through probabilistic dropout of reference images and geometric features during training, along with a controllable CFG scheme during inference. This proposal enables the generation of visually plausible textures, even under scenarios of inconsistency or misalignment between the geometry and the reference image.
The entire framework enables the synthesize of high fidelity, high diversity, and seamless textures for 3D assets as shown in ~\cref{fig:teaser}. We summarize our contributions as follows:
\begin{itemize}
\item[$\bullet$] We introduce 3D-aware Rotary Positional Embedding in our multi-view based texture generation framework, enabling the model to produce highly consistent multi-view images.
\end{itemize}
\begin{itemize}
\item[$\bullet$] We design decoupled multi-attention module, enabling the generated multi-view images that preserve rich 2D diveristy, while simultaneously maintaining superior cross-view consistency and image-following behavior.
\end{itemize} 
\begin{itemize}
\item[$\bullet$] We proposed a geometry-related CFG scheme to balance guidance from both image and geometric conditions, leading to visually plausible texture generation.
\end{itemize}

%% file: srcs/2_relatedwork.tex
\section{Related Work}
\label{sec:related}

\subsection{Diffusion-based Texture Generation.}
\noindent\textbf{Texture generated by Image Space Diffusion.}
Recent advances in texture generation have been inspired by 2D image diffusion models~\cite{rombach2022high} and their multi-view extensions~\cite{shi2023zero123++,shimvdream}. Early methods~\cite{tang2023dreamgaussian,lin2023magic3d,metzer2023latent,poole2022dreamfusion} use score-distillation sampling (SDS) to distill 2D diffusion priors into texture representations, but often result in over-saturated colors and are time-consuming.
Subsequent research has aimed to reduce SDS's computational inefficiency by using geometry-conditioned image diffusion to synthesize multi-view images, followed by texture baking. To maintain multi-view coherence, these methods either use inpainting frameworks~\cite{huang2024material,wu2024texro,zhang2024mapa,ceylan2024matatlas,zeng2024paint3d,chen2023text2tex,richardson2023texture} or synchronization techniques~\cite{liu2025vcd,gao2024genesistex,liu2024text,zhang2024texpainter} to enforce multi-view latent consistency. 
However, inpainting approaches suffer from seams, and synchronized denoising methods often produce plain textures due to loss of latent variations~\cite{liu2025vcd}. Achieving seamless, high-quality textures in multi-view settings remains challenging.

Most existing texture synthesis frameworks, such as Stable Diffusion~\cite{rombach2022high}, rely on text-conditioned generation. However, recent advancements highlight the effectiveness of image-guided approaches, which offer more intuitive guidance, as seen in commercial software like TripoAI~\cite{tripoai2025}, Hunyuan3D~\cite{hunyuan3d2025}, and Rodin~\cite{Hyper3D_AI_2025}. 
Incorporating reference images poses challenges in aligning image and 3D geometry. Solutions like FlexiTex~\cite{jiang2024flexitex} and EASI-Tex~\cite{sairaj_sig24} use IP-Adapter~\cite{ye2023ip-adapter} to inject semantic information, often losing image details. Hunyuan3D-Paint~\cite{zhao2025hunyuan3d} addresses this by integrating image information through a reference branch with a repeating structure from the original diffusion model.
Following this approach, we adopt a reference branch, enhanced with additional disentanglement and 3D-aware positional embedding, to better align geometry and reference images.

\noindent\textbf{Texture generated by 3D Diffusion.}
Research on 3D representation, including UV maps, addresses multi-view image inconsistencies. Methods like Paint3D~\cite{zeng2024paint3d} and Meta3DGen~\cite{bensadoun2024meta} use UV space inpainting, while TexGen~\cite{yu2024texgen} and GeoImageDiffusion~\cite{elizarov2024geometry} train UV-space diffusion models from scratch. Trellis~\cite{xiang2024structured} and TexGaussian~\cite{xiong2024texgaussian} generate 3D Gaussian Splatting (3DGS) representations. 
However, these methods often discard 2D image priors and rely on 3D representations incompatible with 3D engines, limiting their practical use. To inherently avoid the limitations of both directions, our framework introduces a multi-view images synthesis paradigm that seamlessly integrates 2D image priors with explicitly encoded underlying geometry through a 3D-aware positional embedding.

\subsection{Multi-view Diffusion.}
Adapting single-view image diffusion models to multi-view synthesis often results in either multi-view inconsistency or compromised synthesis richness. Multi-view image diffusion has emerged as a promising alternative, enforcing cross-view coherence without sacrificing generative diversity.
Most works modify attention blocks within image diffusion to communicate multi-view latents~\cite{zhao2025hunyuan3d,huang2024mv,vainer2024jointlygeneratingmultiviewconsistent, li2024era3d, tang2025mvdiffusion++, long2024wonder3d, wang2023imagedream, shimvdream, shi2023zero123++, liu2023zero}. For example, Zero123++~\cite{shi2023zero123++} concatenates multi-view latents into a large image and modifies the self-attention block to operate on the entire "multi-view image," correlating latents from all views. Other works inject view constraints into the attention block using diverse attention masks~\cite{Tang2023mvdiffusion, huang2024mv, li2024era3d}.
Acknowledging the superiority of multi-view diffusion, we adopt a similar framework trained on object-centric rendered 3D data~\cite{zhao2025hunyuan3d, objaverse, objaverseXL}.

%% file: srcs/3_method.tex
\section{Methodology}
\label{sec:method}
\begin{figure*}
    \centering
    \includegraphics[width=\linewidth]{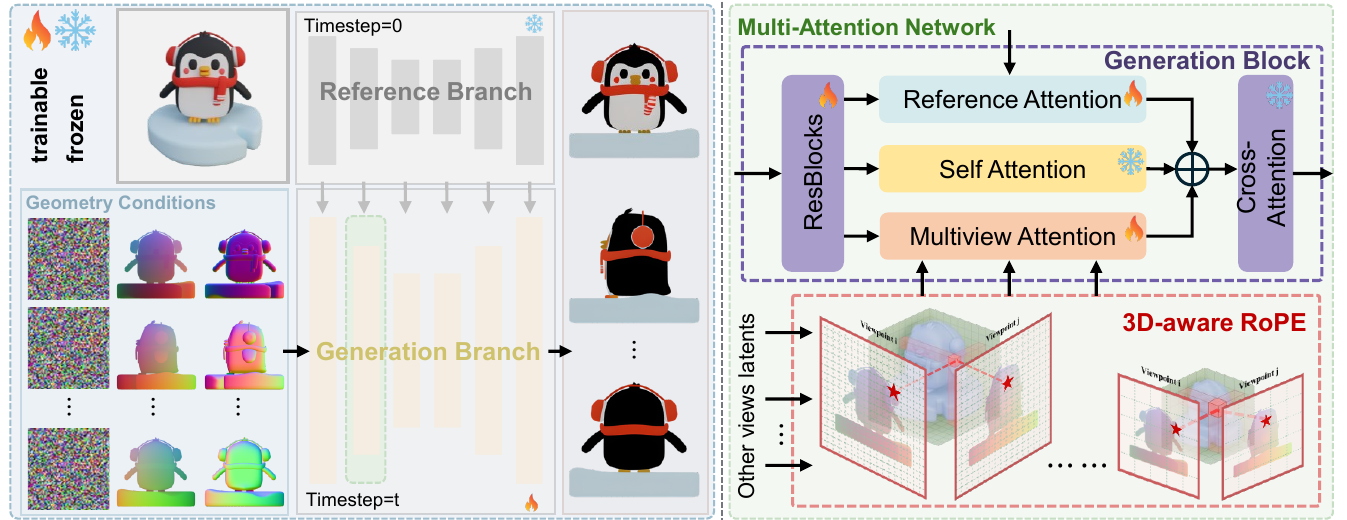}
    \vspace{-5mm}
    \caption{Overview of the proposed texture synthesis framework. Projected geometry conditions and image conditions are incorporated via noise concatenation and reference attention injection, respectively. To enhance multi-view consistency, a multi-view attention block with 3D-aware RoPE is integrated using canonical coordinate maps-based queries.}
    \label{fig:pipeline}
\end{figure*}

As illustrated in Fig.~\ref{fig:pipeline}, our framework aims to generate consistent, high-quality texture maps guided by reference images and 3D mesh structures through a novel multi-view geometry-aware diffusion model. Specifically, we initially extended an image diffusion model (Stable Diffusion~\cite{rombach2022high}) into a image-guided multi-view one by redesigning the self-attention block. 
Specifically, for each self-attention block inside the diffusion model, we augmented it with a reference attention (where the keys and values are obtained using a frozen-weight reference network) and a multi-view attention to increase multi-view consistency.
Following~\cite{zhao2025hunyuan3d, bensadoun2024meta}, we concatenated the Canonical Coordinates Map (CCM) and normal maps in the world coordinate system with the denoised latents for geometry condition injection.
Based on this pipeline, RomanTex innovatively proposed three key novel components designed to address specific challenges in 3D texture synthesis:

\noindent\textbf{(1) 3D-aware Rotary Position Embedding.} The embedding served as the foundation of our work, endowing the generated multi-view images with enhanced consistency and geometry awareness (Sec.~\ref{subsec:rope}). 
By explicitly modeling geometric relationship, achieves seamless textures with continuous appearance while preserving high diversity.

\noindent\textbf{(2) Decoupled Multi-attention Module.} 
Building upon the foundation of~\cite{zhao2025hunyuan3d}, we proposed a training strategy that separates the entanglement between reference injection and multi-view consistency, improving back-view texture quality while maintaining reference fidelity, as demonstrated in Sec.~\ref{subsec:decoupled-ref}.

\noindent\textbf{(3) Geometry-aligned inference.} We proposed an orthogonal Classifier-Free Guidance (CFG) technique (\cref{subsec:geocfg}) that dynamically balances the fidelity to both the reference image and geometry, demonstrating great importance when texturing non-regular geometries (such as generated ones).

\subsection{Preliminary}

\noindent\textbf{Positional Embedding}
The attention mechanism plays a key role in the development of Diffusion models~\cite{esser2024scaling,podell2023sdxl,peebles2023scalable,rombach2022high}, where the main process can be expressed as:
\begin{equation}
    o_{i} = \sum_{j}\alpha_{i,j} \cdot v_j \quad \text{where} \quad \alpha_{i,j} = \mathrm{Softmax}(q_i^T k_j)
\end{equation}
However, a full attention calculation lacks inherent positional awareness~\cite{tang2025mvdiffusion++, shi2023zero123++}, which proves to be suboptimal for spatially-sensitive tasks, such as texture synthesis for a given 3D geometry.
Therefore, Positional embedding (PE) was proposed to address such limitations through positional prior injection. 
Unlike their counterparts with specified attention masks~\cite{huang2024mv, li2024era3d, Tang2023mvdiffusion}, which constrain attention calculation via a specified mask directly, PE implements \textit{soft regularization} by modifying token correlations via:
\begin{equation}
    \alpha'_{i,j} = \mathrm{Softmax}(q_i^T k_j + \phi(i,j))
\end{equation}
where $\phi(i,j)$ encodes positional relationships. 
RoPE~\cite{su2024roformer} establishes a new paradigm through \textbf{rotational position embedding}, achieving unified modeling of absolute and relative positions of specified tokens, which is expressed as:
\begin{equation}
    f(\bm{x}_m, m) = \bm{R}_m \bm{x}_m \quad \text{where} \quad \bm{R}_m = \begin{pmatrix}
    \cos m\theta & -\sin m\theta \\
    \sin m\theta & \cos m\theta
    \end{pmatrix}
\end{equation}
Inspired by these advancements, we devised a 3D-aware Rotary Positional Embedding, which incorporates the given 3D geometry into the framework by the positional embedding, to enhance multi-view consistency for texture generation.

\subsection{3D-aware Rotary Positional Embedding}
\label{subsec:rope}

As illustrated in ~\cref{fig:pipeline}, our framework resolves multi-view inconsistency through a 3D-aware position embedding mechanism that fundamentally differs from prior approaches. While existing methods rely on UV mapping for spatial regularization ~\cite{zhang2024texpainter,liu2024text} or enforce local attention masks ~\cite{liu2025vcd,tang2025mvdiffusion++}, we directly inject 3D structural priors into the latent feature space. This enables multi-view hidden states directly interact with the spatial voxel features of the geometric mesh, thereby integrating native 3D consistency.

Our framework synthesizes multi-view images ${I_i}$ for 3D mesh $M$ through geometrically-aligned latent grids that explicitly encode 2D-3D correspondences. By establishing direct pixel-to-voxel mappings, this mechanism creates cross-view geometric dependencies in the latent space, allowing all perspective-specific generation processes to share a unified geometric foundation, thereby ensuring cross-view consistency in synthesized imagery.

\noindent\textbf{Multi-resolution 2D-3D correspondences. }
Our framework introduces a hierarchical coordinate correspondence strategy to integrate 3D geometry with the multi-scale feature maps within the UNet architecture in Stable Diffusion~\cite{rombach2022high}. 
First, we construct a pyramid of volumetric representations $\{\mathbf{V}^l \in \mathbb{R}^{H^l \times W^l}\}_{l=1}^L$ through progressive downsampling, where each level $l$ corresponds to a specific resolution $r$ of the UNet features. 
For a feature map at a given resolution $r$, we downsample the Canonical Coordinate Map (CCM) to match this resolution, providing the necessary 3D positional information for the feature map. 
Next, we use the obtained 3D positions to query the voxel indices at the corresponding level of the volume. This process can be expressed as:
\begin{equation}
\begin{aligned}
&\left\{
\begin{aligned}
q^{l}(i,j) &= Q f^{l}(i,j) + \phi^{l}(i,j) \\
k^{l}(i,j) &= K f^{l}(i,j) + \phi^{l}(i,j)
\end{aligned}
\right. \\
\end{aligned}
\label{eq:unified}
\end{equation}

\begin{subequations}
\begin{align}
\begin{split}
\phi(i,j) = f_{PE}&\left\{\mathcal{V}^l \left[
\begin{array}{c}
\text{round}\left(pos^{l}_x(i,j) \cdot R^l\right), \\
\text{round}\left(pos^{l}_y(i,j) \cdot R^l\right), \\
\text{round}\left(pos^{l}_z(i,j) \cdot R^l\right)
\end{array}
\right]\right\}
\end{split} \label{eq:phi-def} \\
&pos^{l}(i,j) = \mathrm{CCM}[i,j;\, l]  \label{eq:pos-mapping}
\end{align}
\label{eq:pixel-to-voxel}
\end{subequations}
\noindent where $f^{l}(i,j), \phi^l(i,j), pos^l(i,j)$ represent the $(i,j)$ pixel of l.th-level of feature map, l.th-level positional embedding of $(i,j)$, and l.th-level 3D position, respectively.
This spatial quantization facilitates a bidirectional mapping between 2D feature grids in the image space and their corresponding 3D counterparts. This linkage not only enhances the accuracy of geometric conditioning, achieved by utilizing the original downsampled CCMs without VAE encoding, but also, more importantly, incorporates 3D correspondences into the attention mechanism. This integration leads to improved cross-view consistency.

\noindent\textbf{3D Position Embedding. }
Inspired by Rotary Position Embedding (RoPE)~\cite{su2024roformer} in language modeling, we inject positional information by rotating hidden state vectors according to the corresponding voxel indices $\phi(i,j)$.
The core advancement of 3D-aware RoPE is its ability to provide attention features with additional geometry cues via positional embedding, enhancing the correlation of latents based on 3D proximity. We follow the original RoPE implementation~\cite{su2024roformer} (see supplementary materials for 3D-aware RoPE specifics). When computing the dot product between a query $Q_i$ (at voxel p) and a key $K_j$ (at voxel q), the rotated representations satisfy:
\begin{equation}
Q_i^{rot}\cdot K_j^{rot} \cos(||\theta_p-\theta_q||),    
\end{equation}
where $||\theta_p-\theta_q||$ measures the angular distance between voxels in 3D space. This property ensures local geometric continuity (higher attention for adjacent voxels, smooth decay with distance) and cross-view consistency (maximal similarity for identical 3D regions), aligning feature interactions with 3D spatial priors.

\subsection{Decoupled Multi-attention Module}
\label{subsec:decoupled-ref}
Following Hunyuan3DPaint~\cite{zhao2025hunyuan3d}, we approached the multi-conditioned multi-view images synthesis by a parallel attention structure, extended from the original self-attention block inside the Stable Diffusion~\cite{rombach2022high}. 
To be specific, the parallel structure integrates three distinct blocks, dubbed as \textbf{(1) self attention block (SA),} which we kept the weights frozen to enable our framework with generalization capability; 
\textbf{(2) multi-view attention block (MVA),} which inherits the 3D-aware Rotary Positional Embedding to faciliate coherance among multiple generated images;
\textbf{(3) reference attention block (RefA),} which injects image condition into the Diffusion model.
The parallel structure are implemented in a addition style, written as:

\begin{equation} \label{eq:parallel_structure}
\begin{aligned}[b]
\hat{Z}_{v\in V_i} = &Z_{v\in V_i}+\mathop{\mathrm{Softmax}}\biggl(\frac{QK^T}{\sqrt{d}}\biggr)V + \\
\lambda_{\text{ref}} &\cdot \mathop{\mathrm{Softmax}}\biggl(\frac{Q_{\text{ref}}K_{\text{ref}}^T}{\sqrt{d}}\biggr)V_{\text{ref}} + \\
\lambda_{\text{mv}} &\cdot \mathop{\mathrm{Softmax}}\biggl(\frac{Q_{\text{mv}}K_{\text{mv}}^T}{\sqrt{d}}\biggr)V_{\text{mv}}
\end{aligned}
\end{equation}
where
\begin{subequations} \label{eq:supp_for_parallel_structure}
\begin{align}
&Q, K, V = \{Q,K,V\}^{\text{SA}}_{\text{proj}}(z_{v\in Vi}) \label{eq:def1} \\
&Q_{\text{ref}}, K_{\text{ref}}, V_{\text{ref}} = Q^{\text{RefA}}_{\text{proj}}(z_{v\in Vi}),\,
    \{K,V\}^{\text{RefA}}_{\text{proj}}(Z_{\text{ref}}) \label{eq:def2} \\
&Q_{\text{mv}}, K_{\text{mv}}, V_{\text{mv}} = \{Q,K,V\}^{\text{MVA}}_{\text{proj}}\bigl(z_{v\in \mathcal{V}}+\phi(z_{v\in \mathcal{V}})\bigr) \label{eq:def3}
\end{align}
\end{subequations}

Although specified with distinct tasks for each attention block, the parallel structure often demonstrates functional entangling in practical implementations, particularly pronounced between Multi-View Attention (MVA) and Reference Attention (RefA). 
This architectural coupling induces instability in image generation for camera poses beyond the reference image and simultaneously renders the model extremely prone to classifier-free guidance (CFG) scale.

By examining the framework carfully, we designed a simple yet effective drop-out strategy. Instead of dropping out the reference image solely, as is commonly done to enhance image-following capability by Classifier-free Guidance, we also dropped out the MVA during the training period. This is made possible by the parallel addition structure, where we realized the drop-out of the multi-view branch by setting the $\lambda_{mv}$ in ~\cref{eq:parallel_structure} to zero.

\subsection{Geometry-related Classifier-Free Guidance}
\label{subsec:geocfg}
The consistency between the synthesized texture with the given geometry is also important and has always been the weakness for multi-view-images-based texture synthesis framework compared with original 3D texture generation works~\cite{xiang2024structured, xiong2024texgaussian}.
We approach the challenge by harnessing the power of classifier-free guidance.
Specifically, we apply a CFG operation towards 3D geometry condition, which is represented as a concatnation of the Canonical Coordinate Map (CCM) and the Normal Map.
To make it possible, we also dropped out geometry condition the during training by setting the VAE-latents of the CCM and normal map latents.
Following~\cite{brooks2023instructpix2pix}, we implement the specific multi-conditioned CFG scheme as:
\begin{equation}
\begin{aligned}
    \widetilde{\epsilon_\theta}(z_t,C_{geo},C_{ref}) = & \epsilon_\theta(z_t,\emptyset,\emptyset) \\
    +&\left(\epsilon_\theta(z_t, C_{geo}, \emptyset) - \epsilon_\theta(z_t,\emptyset,\emptyset)\right)\\
    +&\left(\epsilon_\theta(z_t, C_{geo}, C_{ref})- \epsilon_\theta(z_t, C_{geo}, \emptyset)\right)
\end{aligned}
\end{equation}
\noindent where $\epsilon_\theta$ represents the noise estimation, and $\emptyset$ indicates the absence of a specific condition.

While the geometry-conditioned CFG resolves most alignment issues, it sometimes fails when dealing with 3D-generated geometry due to potential misalignment between the geometry and the reference image. To address this conflict between the geometry and reference image conditions, we designed a CFG projection technique. The motivation behind this technique is to follow the geometry by aligning the details when it has sharp activations and to rely on the reference image when the geometry is relatively plain.
To be specific, we modify the last term of CFG equation as:

\begin{equation} \label{eq:epsilon-theta}
\begin{split}
\epsilon_\theta = & \left( \epsilon_\theta(z_t, C_{\text{geo}}, \emptyset) - \epsilon_\theta(z_t, \emptyset, \emptyset) \right) \\
- \Biggl[ &\left( \epsilon_\theta(z_t, C_{\text{geo}}, C_{\text{ref}}) - \epsilon_\theta(z_t, C_{\text{geo}}, \emptyset) \right)^T \\
&\left( \epsilon_\theta(z_t, C_{\text{geo}}, \emptyset) - \epsilon_\theta(z_t, \emptyset, \emptyset) \right) \Biggr]
\end{split}
\end{equation}

\subsection{Implementation Details}
We primarily based our main implementation details and dataset rendering (such as the view selection strategy: 6 for training, 6-12 for inference, and the number of training views) on Hunyuan3D-2.0~\cite{zhao2025hunyuan3d}, due to its proven effectiveness in dense-view inference and rendering dataset construction. For more comprehensive information, please refer to the original report.
Additionally, we randomly drop out the geometry conditions, reference image, and the multi-view attention block (MVA) with three independent probabilities of 0.1 to enhance the flexibility of our work.
We based our model on Stable Diffusion 2~\cite{rombach2022high}, initializing it with the Zero-SNR checkpoint~\cite{lin2024common}, and trained it using the AdamW optimizer with a learning rate of $5 \times 10^{-5}$. 
The training scheme incorporates 2000 warm-up steps and takes around 228 GPU days to complete.

%% file: srcs/4_experiment.tex
\section{Experiment}
\label{sec:experiment}

To systematically evaluate our texture generation framework, we first conducted extensive comparisons with state-of-the-art texture synthesis baselines. Comprehensive qualitative and quantitative analysis demonstrated the superiority of our approach. 
In addition to these comparisons, we conducted extensive user studies to demonstrate that our model is capable of generating texture results that better align with human perceptual preferences.
This was followed by an ablation study examining the contribution of each distinct module by disabling individual components in the framework.

\subsection{Baselines and Metrics}
We compared the proposed method, RomanTex, with comprehensive texture synthesis baselines to demonstrate its effectiveness. 
These baselines include \textit{text-to-texture} approaches:
Text2Tex~\cite{chen2023text2tex}, SyncMVD~\cite{liu2024text}, TexPainter~\cite{zhang2024texpainter}, and Paint3D~\cite{zeng2024paint3d};
\textit{image-to-texture approaches}: Paint3D-IPA~\cite{zeng2024paint3d}, SyncMVD-IPA~\cite{liu2024text};
\textit{text\&image-to-texture} approaches: TexGen~\cite{yu2024texgen}, and HY3D-2.0~\cite{zhao2025hunyuan3d}.
Since most of the existing work focuses on a text-to-texture task, we improved the original SyncMVD~\cite{liu2024text} by incorporating the SDXL-base model~\cite{podell2023sdxl} (which is vital for their framework) and an additional Image IP-adapter~\cite{ye2023ip-adapter} to align with an image-to-texture task and compared it~(referred to as SyncMVD-IPA)  with our approach.
Given the absence of publicly available multi-view image diffusion-based texture generation approaches, with the exception of Hunyuan3D-2.0~\cite{zhao2025hunyuan3d}, our comparisons are necessarily limited to this method.
For TexGen,we utilize partial texture maps generated by our mothed as input, as it requires initial reference images that are completely aligned with input geometries.
For image-to-texture comparisons, we used Jimeng AI~\cite{JM_AI} to generate reference images from randomly generated prompts and employed Hunyuan3D~\cite{hunyuan3d2025} to create the geometry based on these reference images.

For numerical assessment, we adopted criteria from two aspects: distributional similarity between textured mesh renderings and realistic images (FID~\cite{parmar2021cleanfid,heusel2017gans}, CMMD~\cite{jayasumana2024rethinking}), and prompt fidelity (CLIP-T score, CLIP-I score~\cite{radford2021learning}, LPIPS~\cite{zhang2018perceptual}), as is done in HY3D-2.0~\cite{zhao2025hunyuan3d}.
To ensure a fair comparison with existing text-to-texture methodologies, we employ a depth-based ControlNet~\cite{zhang2023adding} to generate an image from a random view of the mesh, which serves as the reference image for our model. 
For more comprehensive details regarding the comparison-baselines and evaluation criteria and experimental setup, please refer to the supplementary materials.

\subsection{Comparisons}
\begin{figure}
    \centering
    \hspace{-3mm}
    \includegraphics[width=1.0\columnwidth]{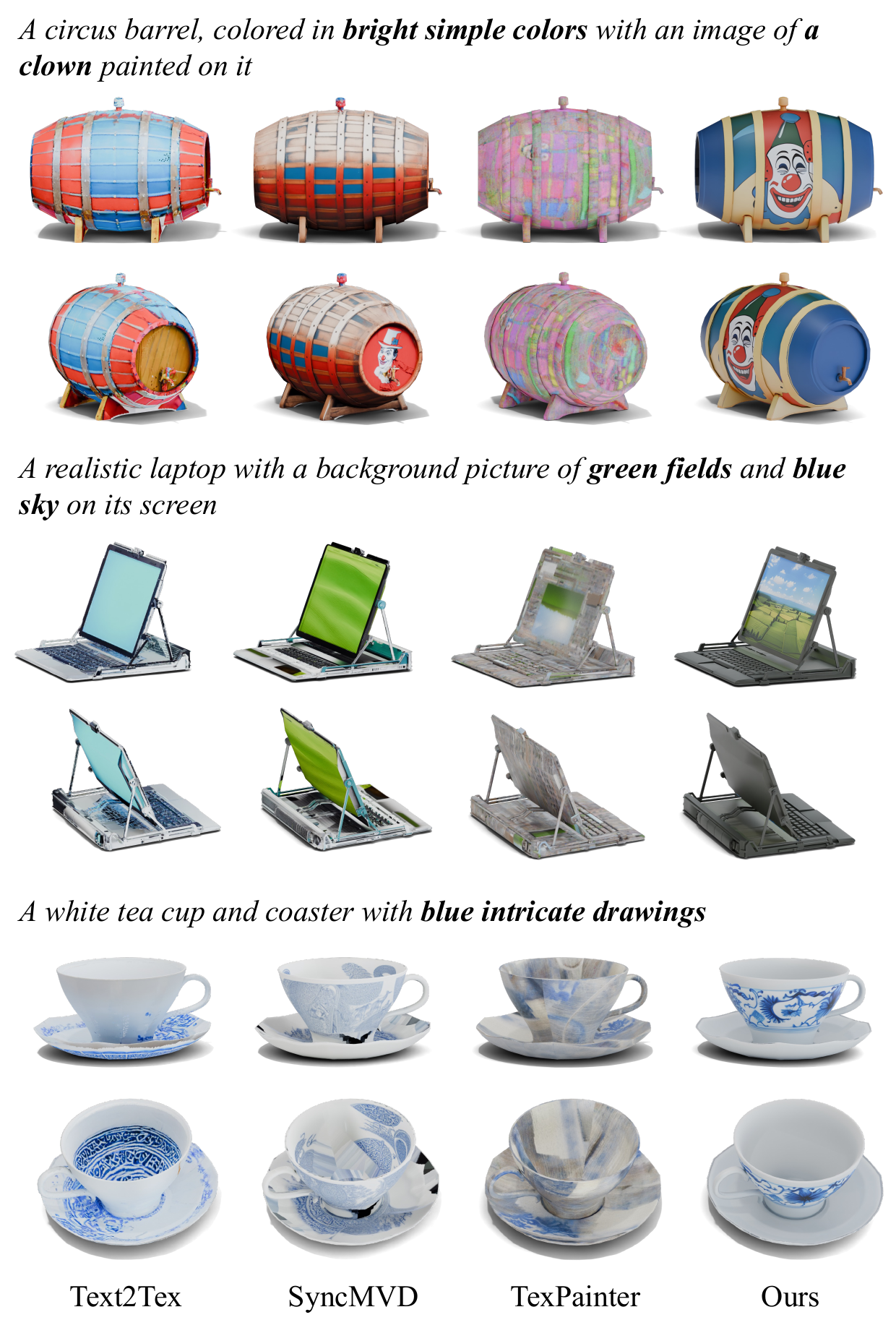}
\vspace{-2mm}
\caption{Visual comparion with text-to-texure methods. We simultaneously present two perspectives to compare consistency performance, and the scheme is also extended to the visual comparison of image-to-texture methods.}
    \label{fig:t2t}
\end{figure}

\begin{figure}
    \centering
    \hspace{-3mm}
    \includegraphics[width=1.0\columnwidth]{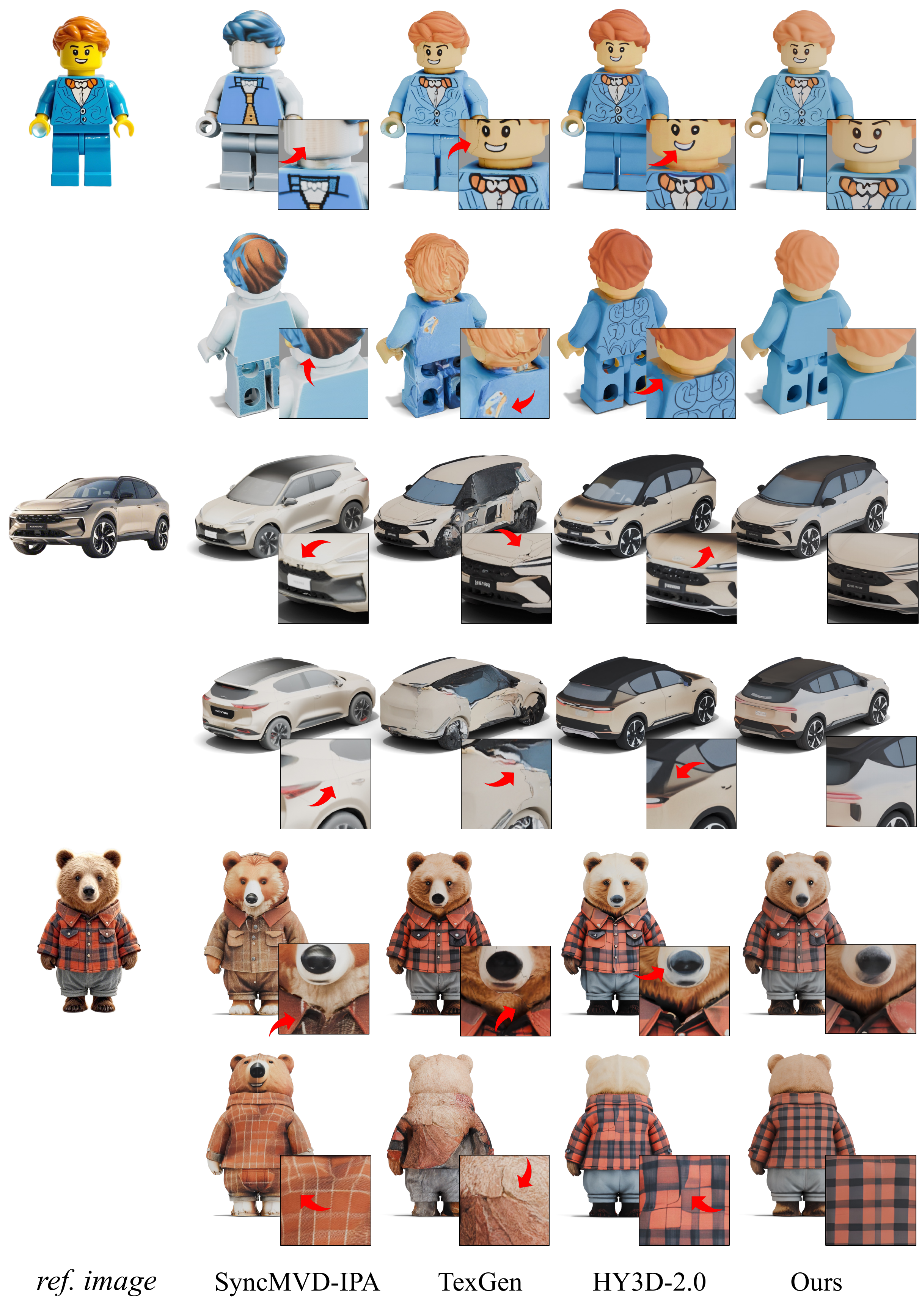}
\vspace{-2mm}
\caption{Visual comparion with image-to-texure methods.We conducted zoomed-in visualization of local regions to enable granular evaluation of detail texture quality.}
    \label{fig:i2t}
\end{figure}

To ensure a fair comparison with the predominantly existing text-guided texture synthesis methods, we adapted our framework to align with them by pre-generating a reference image by adopting a text-to-image pipeline (e.g., Stable Diffusion 1.5 \cite{rombach2022high}) with a depth-based ControlNet \cite{zhang2023adding}. This generated image is then used as a reference input for our image-to-texture process.
As visually and numerically demonstrated in~\cref{fig:t2t},~\cref{fig:i2t} and ~\cref{tab:img2tex}, ~\cref{tab:text2tex}, our RomanTex framework showcases significant improvements over existing methods (both text-to-texture and image-to-texture) in overall texture quality, demonstrating the lowest FiD, CMMD, and enhanced text / image alignment, refering to CLIP-T / CLIP-I and LPIPS (see~\cref{tab:img2tex} and ~\cref{tab:text2tex}). 

Qualitative comparison with three text-to-texture baselines is shown in ~\cref{fig:t2t}.The results of Text2Tex\cite{chen2023text2tex} exhibit a notable deficiency in global consistency, which generates cluttered and disorganized textures; SyncMVD\cite{liu2024text} produces visually more plausible textures but often yields oversmoothed results and unintended gradient artifacts; TexPainter\cite{zhang2024texpainter} frequently introduces noise-corrupted outputs. In contrast, our method achieves texture synthesis that strictly aligns with textual guidance, while generating coherent results in geometrically challenging regions (e.g., back areas as row 4 and geometric self-occlusion area as row 5).

A detailed comparison of image-to-texture task is presented in ~\cref{fig:i2t}. SyncMVD-IPA struggles to synthesize textures with high fidelity, due to its reliance on CLIP's image encoder, which loses fine-grained details of reference images. And it also suffers from the Janus problem. Even using partial textures generated by our method as initialization, TexGen fails to plausibly inpaint missing regions and introduces seam artifacts at UV boundaries. HY3D-2.0~\cite{hunyuan3d2025} produces higher fidelity results, but still suffers from ghosting artifacts caused by multiview inconsistency and color fragmentation due to geometric misalignment. Our model demonstrates superior synthesis quality, meanwhile achieving best image fidelity.

\input{tables/comparisons}

\subsubsection{User Study}
We conducted a user study of image-to-texture methods focused on Q (overall texture qulity) and three subdimension: image following, cross-view consistency and diversity. A test set comprising 50 hand-crafted geometries and 100 generated geometries were utilized for comparison, and the same images generated by T2I models as reference images. 
We invited 3 technical artists and 25 non-experts to conduct quality evaluations. Each participant was randomly assigned 30 sets of generated results and instructed to select the optimal result across four evaluation dimensions. The results obtained are presented in the Table~\ref{tab:userstudy}.

\subsection{Ablation Study}
\begin{figure}
    \centering
    \hspace{-3mm}
    \includegraphics[width=1.0\columnwidth]{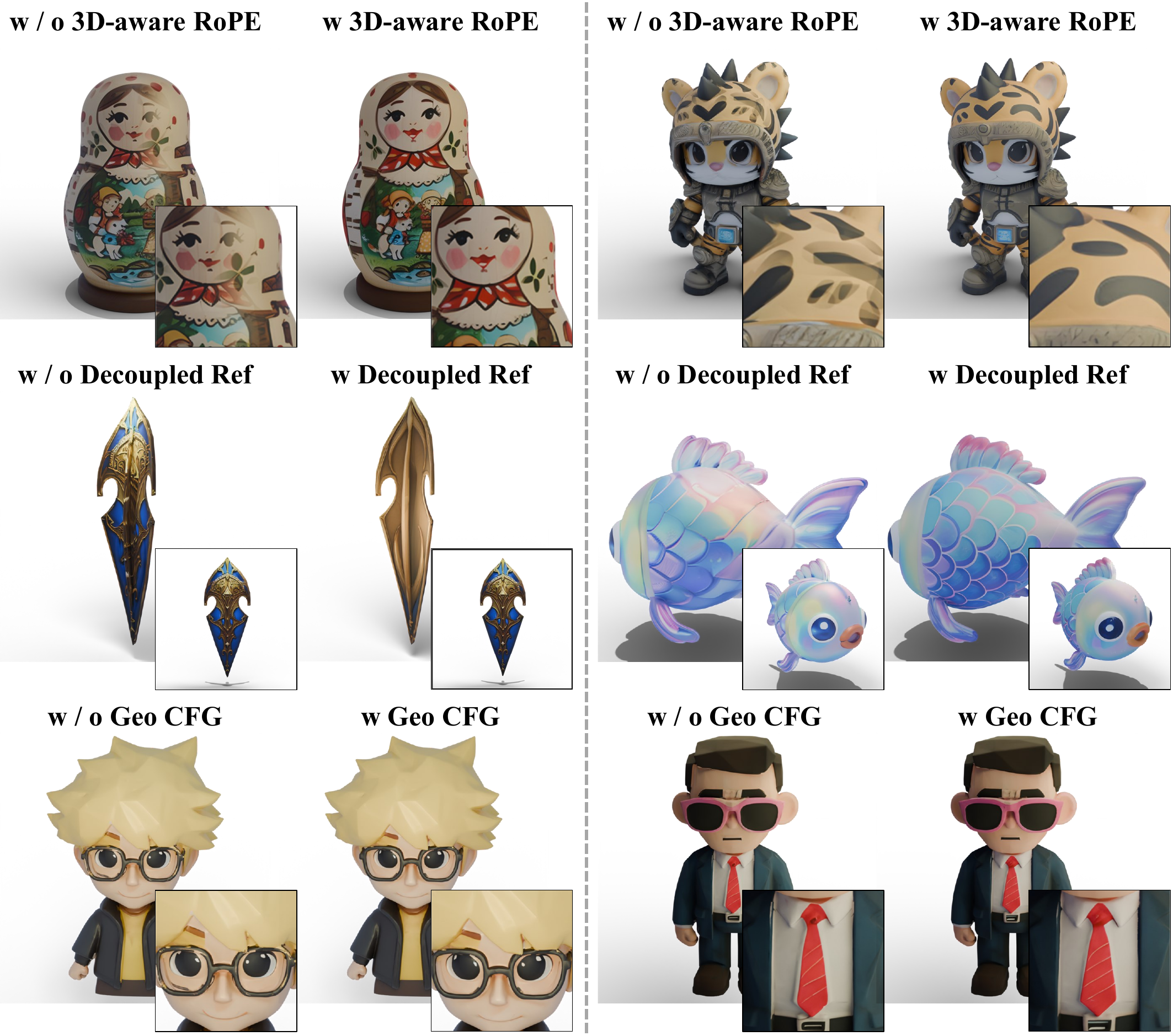}
\vspace{-2mm}
\caption{Ablation study on core components. We validate the effectiveness of our approach by sequentially disabling individual modules: 3D-aware RoPE, Decoupled Reference Branch, and Geometry-related CFG, showcasing their distinct contributions.}
    \label{fig:ablation}
\end{figure}

For Ablation study, specifically, we disabled the 3D-aware RoPE, the decoupled reference branch, and the geometry-related CFG, respectively, to isolate and understand the impact of each component on the overall performance.
These configurations are denoted as "w/o and w 3D-aware RoPE", "w/o and w Decoupled Ref", and "w/o and w Geo CFG".
As illustrated in~\cref{fig:ablation}, the proposed module each faciliates texture generation with multi-view consistency (eliminating ghosting artifacts), back-view regularity (semantic meaningful back-view texture), and accurate geometry-alignment, respectively. 

%% file: tables/comparisons.tex
\definecolor{tabfirst}{RGB}{255,204,204} 
\definecolor{tabsecond}{RGB}{255,229,204} 
\definecolor{tabthird}{RGB}{255,255,204} 

\begin{table}[t]
\centering
\setlength{\tabcolsep}{2.pt}
\renewcommand{\arraystretch}{1.}
\begin{tabular}{l@{\hspace{0.5em}}cccc}
\toprule
\textbf{Method} & {CLIP-FiD $\downarrow$} & {CMMD $\downarrow$} & {CLIP-T$\uparrow$} & {LPIPS$\downarrow$} \\
\midrule
Text2tex~\cite{chen2023text2tex}    & 35.75 & 3.047 & \cellcolor{tabfirst}\bfseries{0.318} & 0.145 \\
SyncMVD~\cite{liu2024text}          & 29.93 & 2.584 & \cellcolor{tabthird}0.307 & 0.141 \\
TexPainter~\cite{zhang2024texpainter} & 28.83 & 2.483 & 0.306 & 0.138 \\
Paint3D~\cite{zeng2024paint3d}      & 30.29 & 2.810 & 0.300 & 0.139 \\
\midrule
TexGen~\cite{yu2024texgen}          & \cellcolor{tabthird}28.24 & \cellcolor{tabthird}2.447 & 0.302 & \cellcolor{tabsecond}0.133 \\
HY3D-2.0~\cite{zhao2025hunyuan3d} & \cellcolor{tabsecond}26.44 & \cellcolor{tabsecond}2.318 & \cellcolor{tabthird}0.307 & \cellcolor{tabthird}0.136 \\
Ours                                & \cellcolor{tabfirst}\bfseries{24.78} & \cellcolor{tabfirst}\bfseries{2.191} & \cellcolor{tabsecond}0.308 & \cellcolor{tabfirst}\bfseries{0.121} \\
\bottomrule
\end{tabular}
\caption{Texture generation criteria showing performance on text-to-texture tasks. 
  Colored cells with \textbf{bold text} indicate best (\colorbox{tabfirst}{\phantom{AA}}), second-best (\colorbox{tabsecond}{\phantom{AA}}), 
  and third-best (\colorbox{tabthird}{\phantom{AA}}) results per column. The following tables are marked in the same way.}
\label{tab:text2tex}
\end{table}

\begin{table}[t]
\centering
\setlength{\tabcolsep}{1.5pt}
\renewcommand{\arraystretch}{1.}
\begin{tabular}{l@{\hspace{0.0em}}cccc}
\toprule
\textbf{Method} & {CLIP-FiD $\downarrow$} & {CMMD $\downarrow$} & {CLIP-I$\uparrow$} & {LPIPS$\downarrow$} \\
\midrule
Paint3D-IPA~\cite{zeng2024paint3d}     & \cellcolor{tabthird}26.86 & 2.400 & \cellcolor{tabfirst}\bfseries{0.998} & \cellcolor{tabsecond}0.126 \\
SyncMVD-IPA~\cite{liu2024text}     & 28.39 & \cellcolor{tabthird}2.397 & 0.882 & 0.142 \\
TexGen~\cite{yu2024texgen}         & 28.237 & 2.448 & 0.867 & 0.133 \\
HY3D-2.0~\cite{zhao2025hunyuan3d} & \cellcolor{tabsecond}26.439 & \cellcolor{tabsecond}2.318 & \cellcolor{tabthird}0.889 & \cellcolor{tabsecond}0.126 \\
Ours                                & \cellcolor{tabfirst}\bfseries{24.78} & \cellcolor{tabfirst}\bfseries{2.191} & \cellcolor{tabsecond}0.891 & \cellcolor{tabfirst}\bfseries{0.121} \\
\bottomrule
\end{tabular}
\caption{Texture generation criteria showing performance on image-to-texture tasks.}
\label{tab:img2tex}
\end{table}

\begin{table}[t]
\centering
\setlength{\tabcolsep}{1.5pt}
\renewcommand{\arraystretch}{1.}
\begin{tabular}{l@{\hspace{0.0em}}cccc}
\toprule
\textbf{Method} & {F(\%) $\uparrow$} & {C(\%) $\uparrow$} & {D(\%) $\uparrow$} & {Q(\%) $\uparrow$} \\
\midrule
SyncMVD-IPA~\cite{liu2024text}     & 3.5 & \cellcolor{tabthird}11.2 & \cellcolor{tabthird}15.6 & \cellcolor{tabthird}10.1 \\
TexGen~\cite{yu2024texgen}         & \cellcolor{tabthird}6.8 & 2.9 & 4.1 & 5.3 \\
HY3D-2.0~\cite{zhao2025hunyuan3d} & \cellcolor{tabsecond}27.6 & \cellcolor{tabsecond}21.5 & \cellcolor{tabsecond}23.3 & \cellcolor{tabsecond}20.7 \\
Ours                                & \cellcolor{tabfirst}\bfseries{62.1} & \cellcolor{tabfirst}\bfseries{64.4} & \cellcolor{tabfirst}\bfseries{57.0} & \cellcolor{tabfirst}\bfseries{63.9} \\
\bottomrule
\end{tabular}
\caption{User study statistics of image-to-texture method. The four indicators are: image following (F), cross-view consistency (C), diversity (D), and overall texture qulity (Q).}
\label{tab:userstudy}
\end{table}

%% file: srcs/5_conclusion.tex
\section{Conclusion}
\label{sec:conclusion}
In this paper, we propose a novel approach for high-quality and consistent texture synthesis. To achieve this, we adapt the multi-view image generation framework with three key enhancements:
We introduce a novel 3D-aware Rotary Positional Embedding into the multi-view attention mechanism within the diffusion model. This significantly improves multi-view consistency through enhanced geometry awareness.
To achieve more reasonable back-view synthesis, we devise a decoupled reference branch using a simple yet effective drop-out training strategy.
To balance image alignment and geometry alignment, we employ a geometry-related Classifier-Free Guidance (CFG) technique during inference.
We demonstrate the effectiveness of our approach through extensive comparisons with state-of-the-art methods and ablation studies on distinct modules, showing fewer seams and artifacts in the synthesized textures. Nevertheless, the proposed work still faces the challenge of baked-in illumination in the generated textures. This issue could be further addressed by extending our pipeline to incorporate physically-based rendering textures, which will be the main focus of our future work.

%% file: srcs/appendix.tex
\section{Supplementary on Baselines and Experimental Criteria}
\label{sec:}
Text2Tex is an inpainting-based approach, while SyncMVD and TexPainter leverage synchronizing techniques (at latent space and image space) to enhance multi-view consistency. 
In contrast, Paint3D employs an inpainting-dependent texture synthesis in the first stage, followed by a UV-based refinement to enhance the texture quality. 
On the other hand, TexGen utilizes an end-to-end UV space diffusion approach to generate textures. However, since their work relies on a incomplete UV texture as a starting point, we used a partial texture baked from the original reference image for initialization.
Since most of the existing work focuses on a text-to-texture task, we improved the original SyncMVD~\cite{liu2024text}~(referred to as SyncMVD-IPA) by incorporating the SDXL-base model~\cite{podell2023sdxl} and an additional Image IP-adapter~\cite{ye2023ip-adapter} to align with an image-to-texture task and compared it with our approach.
Specifically, leveraging the Clean-FID~\cite{parmar2021cleanfid} implementation, we harnessed a CLIP-version of Fréchet Inception Distance $FID_{CLIP}$ to compute the distance re-renderings and the ground-truth renderings.
Besides, the recently proposed CLIP Maximum-Mean Discrepancy (CMMD)~\cite{jayasumana2024rethinking} is also utilized to serve as an complementary criteria to validate the distribution of generated texture. 
In addition to these two metrics, we also use CLIP-I / CLIP-T score~\cite{radford2021learning} to validate semantic alignment between renderings of the generated texture map and given image / text and LPIPS~\cite{zhang2018perceptual} to estimate the consistency between renderings of the generated texture map and the reference images.

\section{Supplementary on Ablation Study}
To further evaluate the effectiveness of 3D-aware RoPE numerically, we introduced a criterion called the local alignment distance (LAD). This score computes the average Mean Squared Error (MSE) over overlapping regions between adjacent views, providing a quantitative measure of multiview coherence. The LAD is defined as follows:
\begin{equation}
    \mathrm{LAD} = \sum \left|\left|M_{v}^{UV}\odot \left\{T_{v}^{UV} - \left[\frac{1}{|\mathcal{V}|}\sum_{v\in \mathcal{V}}T^{UV}_{v}\odot M^{UV}_{v}\right]\right\}\right|\right|^2
\end{equation}
where $T_{v}^{UV}$ and $M_{v}^{UV}$ represent the texture and mask in UV space unwrapped from the image of view $v$.
\input{tables/ablation}
As demonstrated in ~\cref{tab:LAD}, our 3D-aware RoPE significantly outperforms the naive self-attention mechanism that lacks 3D geometry awareness.

\section{Supplementary on 3D-aware RoPE}
Inspired by Rotary Position Embedding (RoPE)~\cite{su2024roformer} in language modeling, we inject positional information by rotating hidden state vectors according to the corresponding voxels, expressed as
\begin{equation}
\setlength{\arraycolsep}{2pt}
\renewcommand{\arraystretch}{0.7}
\begin{split}
f(x_m, (x,y,z)) &= 
\begin{pmatrix}
x_0 \\ x_1 \\ \vdots \\ x_m \\ x_{m+1} \\ \vdots \\ x_{2m+1} \\ x_{2m+2} \\ \vdots \\ x_{3m+3}
\end{pmatrix} 
\otimes
\begin{pmatrix}
\cos x\theta_0 \\ \cos x\theta_0 \\ \vdots \\ \cos x\theta_{\lfloor m/2 \rfloor} \\ 
\cos y\theta_{\lfloor (m+1)/2 \rfloor} \\ \vdots \\ \cos y\theta_m \\ 
\cos z\theta_{m+1} \\ \vdots \\ \cos z\theta_{\lfloor (3m+3)/2 \rfloor}
\end{pmatrix} 
+\\
&\quad
\begin{pmatrix}
x_1 \\ -x_0 \\ \vdots \\ -x_{m-1} \\ x_{m+2} \\ \vdots \\ -x_{2m} \\ 
x_{2m+3} \\ \vdots \\ x_{3m+2}
\end{pmatrix} 
\otimes
\begin{pmatrix}
\sin x\theta_0 \\ \sin x\theta_0 \\ \vdots \\ \sin x\theta_{\lfloor m/2 \rfloor} \\ 
\sin y\theta_{\lfloor (m+1)/2 \rfloor} \\ \vdots \\ \sin y\theta_m \\ 
\sin z\theta_{m+1} \\ \vdots \\ \sin z\theta_{\lfloor (3m+3)/2 \rfloor}
\end{pmatrix}
\end{split}
\end{equation}

%% file: tables/ablation.tex
\begin{table}[htbp]
\centering

\label{tab:results}
\scalebox{0.9}{
\begin{tabular}{l@{\hspace{2em}}c}
\toprule
\multicolumn{1}{c}{\textbf{Method}} & \textbf{LAD} \\
\midrule
\textbf{w\slash o} MVA               & 0.142 \\ 
\textbf{w\slash o} 3D-aware RoPE     & 0.123 \\
\textbf{w\slash o} 3D-aware RoPE     & \textbf{0.119} \\
\bottomrule
\label{tab:LAD}
\end{tabular}
}
\vspace{-3pt}
\caption{Ablation study on local alignment distance (LAD)}
\vspace*{-8pt}
\end{table}